\documentclass{article}
\usepackage{spconf,amsmath,graphicx}

\usepackage{enumitem}
\setlist{nosep, leftmargin=14pt}

\usepackage{mwe} 
\usepackage{graphicx}    
\usepackage{amsmath}     
\usepackage{algorithm}   
\usepackage{algpseudocode}
\usepackage{subfigure}   
\usepackage{booktabs}    
\usepackage{xcolor}      
\usepackage{amsfonts}    
\usepackage{nicefrac}    
\usepackage{url}         
\usepackage{wrapfig}     
\usepackage{multirow}    
\usepackage{newunicodechar} 
\usepackage[capitalize,noabbrev]{cleveref} 
\usepackage{enumitem}
\usepackage{multicol}
\usepackage{adjustbox}
\usepackage{float}
\usepackage{adjustbox}
\newcommand{\header}[1]{\noindent\textbf{{#1.}}~~}

\usepackage[utf8]{inputenc} 
\usepackage[T1]{fontenc}    
\usepackage{url}            
\usepackage{booktabs}       
\usepackage{amsfonts}       
\usepackage{nicefrac}       
\usepackage{microtype}      
\usepackage{xcolor}         

\title{Adaptive Clinical-Aware Latent Diffusion for Multimodal Brain Image Generation and Missing Modality Imputation}

\name{\parbox{2\columnwidth}{\centering Rong Zhou$^{1}$, Houliang Zhou$^{1}$, Yao Su$^{2}$, Brian Y. Chen$^{1}$, Yu Zhang$^{3,4,5}$, Lifang He$^{1}$, \\Alzheimer's Disease Neuroimaging Initiative$^{6}$}}
\address{$^{1}$Department of Computer Science and Engineering, Lehigh University, PA, USA\\
$^{2}$Worcester Polytechnic Institute, MA, USA\\
$^{3}$Department of Psychiatry and Behavioral Sciences, Stanford University School of Medicine, CA, USA\\
$^{4}$Wu Tsai Neurosciences Institute, Stanford University, CA, USA\\
$^{5}$Stanford Institute for Human-Centered AI, Stanford, CA, USA\\
$^{6}$Alzheimer's Disease Neuroimaging Initiative
}
\begin{document}

\maketitle
\begin{abstract}
Multimodal neuroimaging provides complementary insights for Alzheimer's disease diagnosis, yet clinical datasets frequently suffer from missing modalities. We propose ACADiff, a framework that synthesizes missing brain imaging modalities through adaptive clinical-aware diffusion. ACADiff learns mappings between incomplete multimodal observations and target modalities by progressively denoising latent representations while attending to available imaging data and clinical metadata. The framework employs adaptive fusion that dynamically reconfigures based on input availability, coupled with semantic clinical guidance via GPT-4o-encoded prompts. Three specialized generators enable bidirectional synthesis among sMRI, FDG-PET, and AV45-PET. Evaluated on ADNI subjects, ACADiff achieves superior generation quality and maintains robust diagnostic performance even under extreme 80\% missing scenarios, outperforming all existing baselines. To promote reproducibility, code is available at \url{https://github.com/rongzhou7/ACADiff}.
\end{abstract}
\begin{keywords}
Latent diffusion, multimodal imaging, missing modality imputation, Alzheimer's disease
\end{keywords}
\section{Introduction}
\label{sec.intro}
Multimodal neuroimaging has become increasingly essential for understanding Alzheimer's disease (AD), as different modalities capture complementary pathological aspects~\cite{jack2010hypothetical}. For example, MRI quantifies structural brain atrophy, FDG-PET measures regional glucose metabolism, and AV45-PET reveals amyloid deposition~\cite{jagust2018imaging}. Together, these modalities offer a more complete characterization of the underlying disease process compared to any single modality alone~\cite{xu2024comprehensive}.

Unfortunately, this potential is limited in part because real-world datasets frequently suffer from incomplete modalities, where not all imaging scans are available for every subject due to prohibitive cost, acquisition protocol variability, or unexpected patient dropout \cite{mueller2005alzheimer}. This incompleteness limits both research insights and clinical decision-making.

Recent advances in generative modeling have shown promise for synthesizing missing modalities~\cite{eigenschink2023deep}. While conditional GANs like Pix2Pix~\cite{isola2017image} and DS-GAN~\cite{pan2019disease} provide initial solutions, they often suffer from mode collapse and training instability. Recent diffusion models~\cite{rombach2022high,yu2024functional,li2024pasta} offer improved stability and generation quality. However, existing approaches still face key limitations: (1) lack of adaptive fusion mechanisms for varying input combinations; (2) limited integration of clinical information beyond disease labels; (3) absence of semantic understanding of medical metadata.

\begin{figure*}[!t]
	\centering
   \includegraphics[width=0.9\linewidth]{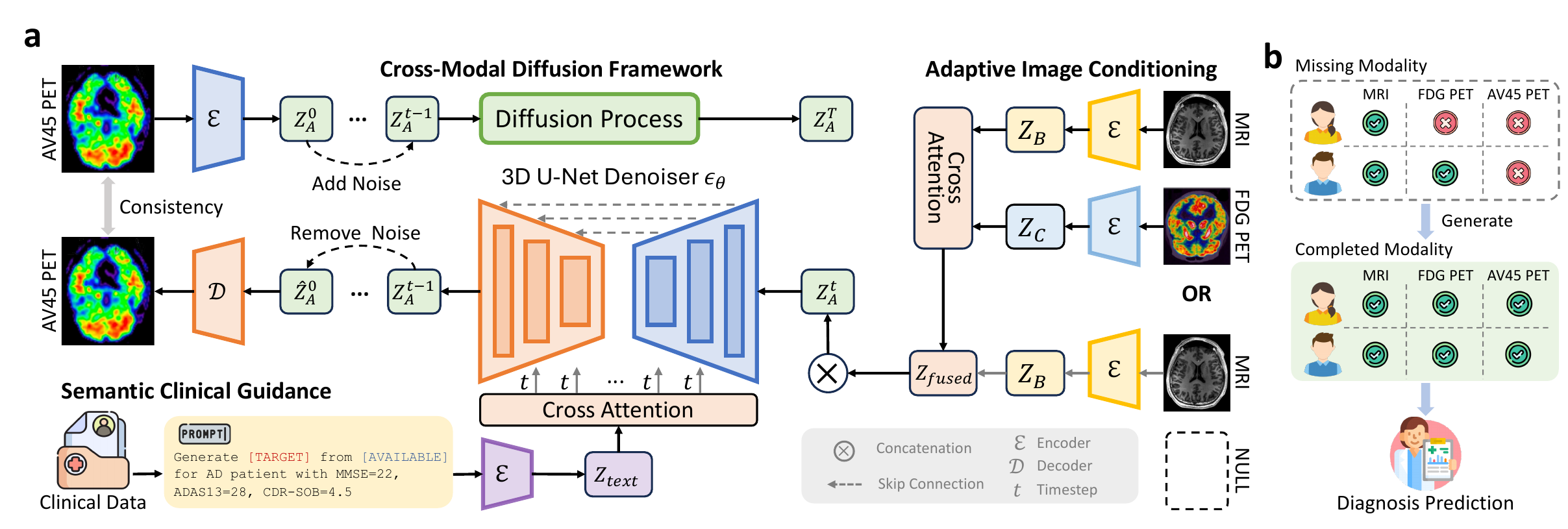}
   \vspace{-10pt}
    \caption{Overview of ACADiff. \textbf{(a)} Cross-modal latent diffusion with adaptive multi-source image conditioning and semantic clinical guidance via GPT-4o-encoded prompts. \textbf{(b)} Missing modality imputation for downstream diagnosis.}
\label{fig:framework}
\vspace{-15pt}
\end{figure*}

To address these challenges, we propose ACADiff (Adaptive Clinical-Aware Diffusion), a framework that synthesizes missing brain imaging modalities through hierarchical conditional diffusion. ACADiff learns the complex mapping between incomplete multimodal observations and target modalities by progressively denoising latent representations while simultaneously attending to available imaging data and clinical metadata. The framework employs adaptive fusion strategies that dynamically reconfigure based on input availability, coupled with semantic clinical guidance that ensures disease-relevant patterns are preserved throughout the generation process. Our key contributions are:
\begin{itemize}
\item \textbf{Adaptive multi-source fusion}: A single model seamlessly handles both 2→1 and 1→1 generation through dynamic conditioning that switches between cross-attention for multiple inputs and projection for single modality.

\item \textbf{Clinical-aware synthesis}: Integration of disease labels and continuous cognitive scores (MMSE, ADAS13, CDR-SOB) to guide generation, ensuring synthesized images preserve diagnostic-relevant patterns.

\item \textbf{Comprehensive bidirectional synthesis}: Three specialized generators enable all six translation directions among MRI, FDG-PET, and AV45-PET, each optimized for target-specific characteristics.

\item \textbf{Language model enhancement}: GPT-4o encodes clinical data as structured prompts, providing semantic understanding beyond traditional embedding approaches.
\end{itemize}

Experiments on ADNI subjects demonstrate that ACADiff achieves superior generation quality and maintains robust diagnostic performance across all missing rates. Even with 80\% missing data, our method consistently outperforms existing approaches, validating its clinical utility for multimodal brain image completion. These results establish ACADiff as a robust solution for clinical multimodal completion.

\begin{figure*}[!t]
\centering
\includegraphics[width=\linewidth]{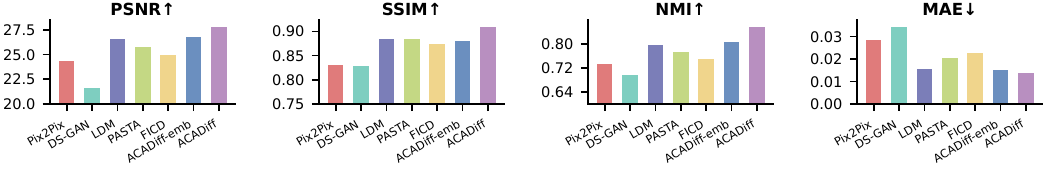}
\vspace{-25pt}
\caption{Image generation performance across methods. Higher PSNR/SSIM/NMI and lower MAE indicate better quality.}
\label{fig:gen}
\vspace{-10pt}
\end{figure*}

\section{Methods}
\label{sec:methods}
Fig.~\ref{fig:framework} illustrates our ACADiff framework, which employs latent diffusion to synthesize missing brain imaging modalities. The model learns to generate target modalities conditioned on available inputs through adaptive fusion and hierarchical conditioning, handling both 2→1 and 1→1 generation scenarios.

\header{Latent Space Construction}Training diffusion models directly on 3D brain imaging is computationally demanding. We employ modality-specific 3D VAEs to map each modality $X_M$ to compact latent representations $Z_M$ through encoders $\boldsymbol{\mathcal{E}}_M$, with paired decoders $\boldsymbol{\mathcal{D}}_M$ for reconstruction.

\header{Cross-Modal Diffusion}We employ denoising diffusion in the latent space to generate target modalities conditioned on available information. The forward process gradually perturbs a clean latent $Z_M^0$ into Gaussian noise $Z_M^T$ over $T$ timesteps. The reverse denoising process is learned as:
\vspace{-5pt}
\[
p_\theta(Z_M^{t-1}|Z_M^t,Z_{\neg M},z_{\text{text}},z_{\text{avail}},t)
=\mathcal{N}(\mu_\theta,\sigma_t^2\mathbf{I}),
\]
where $Z_{\neg M}$ denotes the latent representations of available non-target modalities (e.g., when generating $Z_C$, this could be $\{Z_A, Z_B\}$ for 2→1 generation or just $\{Z_A\}$ for 1→1 generation), $z_{\text{text}}$ is the encoded text embedding, $z_{\text{avail}} \in \{0,1\}^3$ is a binary vector indicating the availability of each modality (e.g., $[1,1,0]$ means modalities
A and B are available while C is missing, and the mean $\mu_\theta$ is parameterized using a 3D U-Net denoiser $\epsilon_\theta$:
\vspace{-5pt}
\[
\mu_\theta=\tfrac{1}{\sqrt{\alpha_t}}\!\left(
Z_M^t-\tfrac{\beta_t}{\sqrt{1-\bar{\alpha}_t}}
\epsilon_\theta(Z_M^t,Z_{\neg M},z_{\text{text}},z_{\text{avail}},t)
\right).\]
We first minimize the noise prediction error:
\vspace{-5pt}
\[
\mathcal{L}_{\text{diff}}=
\mathbb{E}\!\left[
\|\epsilon-\epsilon_\theta(Z_M^t,Z_{\neg M},z_{\text{text}},z_{\text{avail}},t)\|^2
\right].
\]
To ensure generated modalities align with real distributions, we incorporate a consistency regularization: $\mathcal{L}_{\text{cons}} = \|\hat{Z}_M^0 - Z_M^0\|$. The final training objective becomes: $\mathcal{L} = \mathcal{L}_{\text{diff}} + \lambda \mathcal{L}_{\text{cons}}$, where $\lambda$ balances denoising and reconstruction objectives.

\header{Hierarchical Adaptive Conditioning}
The diffusion model integrates three complementary conditions that work hierarchically to enable adaptive generation:

\textit{(1) Adaptive Image Conditioning.}
Available modalities $Z_{\neg M}$ are first fused based on their availability pattern:
\vspace{-5pt}
\[
Z_{\text{fused}} = \begin{cases}
\textit{CrossAttn}(Z_i, Z_j), & \text{if } \sum z_{\text{avail}} = 2 \\
\textit{Proj}(Z_i), & \text{if } \sum z_{\text{avail}} = 1
\end{cases}
\]
where $Z_i, Z_j \in Z_{\neg M}$ denote available modalities. \textit{CrossAttn} 
applies multi-head attention between spatially-pooled features to enable 
inter-modal information exchange, while \textit{Proj} performs learnable 
3D convolution for single-modality adaptation. The fused features are 
concatenated with the noisy target latent $Z_M^t$ as input to the U-Net 
denoiser: $\epsilon_\theta([Z_M^t; Z_{\text{fused}}], t, z_{\text{text}})$, 
where $[;]$ denotes channel-wise concatenation, enabling early fusion of 
cross-modal information.

\textit{(2) Semantic Clinical Guidance.}
Clinical data including disease diagnosis and cognitive scores 
(MMSE, ADAS13, CDR-SOB) are composed into structured prompts: 
"\textit{Generate [TARGET] from [AVAILABLE] for AD patient with MMSE=22, ADAS13=28, CDR-SOB=4.5}". These prompts are encoded by a pretrained 
language model: $z_{\text{text}} = \mathcal{E}(\text{prompt})$, then fused 
into the decoder through cross-attention: 
$F^l \leftarrow F^l + g^l(\textit{CrossAttn}(F^l, z_{\text{text}}))$, 
enabling semantic guidance that aligns generation 
with disease-specific patterns.

\textit{(3) Temporal Modulation}
The timestep $t \in \{0, 1, ..., T\}$ indicates the current noise level during denoising, where larger $t$ corresponds to noisier inputs requiring stronger denoising. The timestep controls denoising dynamics across all layers: $F^l \leftarrow \gamma(t)F^l+\beta(t)$, enabling the model to adaptively adjust its denoising strength at each diffusion step.


\header{Training and Inference} During training, modality dropout creates diverse scenarios: each sample randomly uses either two modalities (2→1) or one modality (1→1) to generate the target. At inference, the model adapts via $z_{\text{avail}}$, applying cross-modal attention for 2→1 or projection for 1→1. After iterative denoising, the decoder produces full-resolution output: $\hat{X}_M = D_M(\hat{Z}_M^0)$. We randomly drop clinical information, enabling generation without clinical guidance when needed.

\section{Experiments and Results}
\header{Data Acquisition and Preprocessing} We utilized 1,028 subjects from the ADNI cohort~\cite{mueller2005alzheimer}, including 198 AD, 495 MCI, and 335 HC with sMRI, FDG-PET, and AV45-PET. sMRI underwent skull stripping, intensity normalization, and nonlinear registration to MNI space. PET scans were co-registered to MRI using the same transformation. All volumes were cropped to $160{\times}180{\times}160$ voxels and normalized to $[-1,1]$.

To avoid data leakage, we split the 1,028 subjects into two independent sets: 600 for generator training (10\% validation, 10\% test). and 428 for classification experiments. From the classification set (428 subjects), we reserved 128 as a held-out test set and used the remaining 300 for classifier training (10\% for validation). 
For classifier training, we simulated missing-modality scenarios where 
20\%, 40\%, 60\%, or 80\% of the 300 training subjects had 1-2 randomly 
selected modalities removed, then imputed using the trained generators.

\header{Experimental Settings}
For generation, we evaluate voxel-level synthesis using MAE, PSNR, SSIM, and NMI within a brain mask. For classification, we measure Accuracy (ACC), Sensitivity (SEN), Specificity (SPE), and AUC.
Five methods are adapted for comparison. (1) \textbf{Pix2Pix}~\cite{isola2017image}: A conditional GAN with 3D U-Net generator trained with adversarial and $L_1$ losses, extended to handle multi-channel concatenation of available modalities. (2) \textbf{DS-GAN}~\cite{pan2019disease}: A disease-aware GAN that incorporates disease labels as auxiliary conditioning and employs spectral normalization to stabilize training. (3) \textbf{LDM}~\cite{rombach2022high}: A latent diffusion model that encodes images into the same compact latent space ($20 \times 22 \times 20$) as our method but uses standard concatenation for multi-modal conditioning without adaptive fusion. (4) \textbf{PASTA}~\cite{li2024pasta}: A pathology-aware diffusion model with dual-arm architecture and cycle consistency, modified to handle multi-source inputs. (5) \textbf{FICD}~\cite{yu2024functional}: A constrained diffusion model with voxel-wise functional alignment losses for metabolic consistency.

\begin{table}[H]
  \centering
  \vspace{-15pt}
  \caption{Classification performance for AD vs. HC under different missing data strategies.}
  \resizebox{\linewidth}{!}{
    \begin{tabular}{cccccc}
    \toprule
    Missing Rate & Method & ACC   & AUC   & SEN   & SPE \\
    \midrule
    0\%  & Oracle (Real) & 0.920±0.029 & 0.943±0.023 & 0.889±0.031 & 0.908±0.028 \\
    \midrule
    \multirow{9}[3]{*}{20\%} 
          & Drop & 0.825±0.042 & 0.852±0.040 & 0.788±0.043 & 0.806±0.041 \\
          & Mean & 0.798±0.045 & 0.826±0.043 & 0.761±0.046 & 0.779±0.044 \\
    \cmidrule{2-6}
          & Pix2Pix & 0.865±0.035 & 0.893±0.032 & 0.821±0.035 & 0.848±0.035 \\
          & DS-GAN & 0.851±0.039 & 0.882±0.037 & 0.805±0.039 & 0.786±0.038 \\
          & LDM   & 0.885±0.036 & 0.902±0.031 & 0.817±0.033 & 0.861±0.038 \\
          & PASTA & 0.883±0.036 & 0.900±0.032 & 0.820±0.035 & 0.850±0.031 \\
          & FICD & 0.859±0.040 & 0.898±0.031 & 0.807±0.037 & 0.848±0.040 \\
          & ACADiff-emb(ours) & 0.891±0.032 & 0.904±0.031 & 0.825±0.033 & 0.865±0.036 \\
          & ACADiff (ours) & \textbf{0.894±0.035} & \textbf{0.910±0.026} & \textbf{0.827±0.034} & \textbf{0.868±0.031} \\
    \midrule
    \multirow{9}[3]{*}{40\%} 
          & Drop & 0.768±0.048 & 0.795±0.046 & 0.731±0.049 & 0.749±0.047 \\
          & Mean & 0.742±0.050 & 0.769±0.048 & 0.705±0.051 & 0.723±0.049 \\
    \cmidrule{2-6}
          & Pix2Pix & 0.853±0.039 & 0.882±0.038 & 0.807±0.040 & 0.841±0.039 \\
          & DS-GAN & 0.847±0.036 & 0.874±0.037 & 0.794±0.038 & 0.776±0.037 \\
          & LDM   & 0.877±0.032 & 0.892±0.032 & 0.809±0.037 & 0.850±0.036 \\
          & PASTA & 0.875±0.033 & 0.897±0.031 & 0.815±0.035 & 0.848±0.034 \\
          & FICD & 0.858±0.036 & 0.888±0.032 & 0.798±0.035 & 0.841±0.034 \\
          & ACADiff-emb(ours) & 0.886±0.031 & 0.902±0.030 & 0.821±0.037 & 0.851±0.035 \\
          & ACADiff (ours) & \textbf{0.889±0.031} & \textbf{0.906±0.025} & \textbf{0.823±0.032} & \textbf{0.854±0.037} \\
    \midrule
    \multirow{9}[4]{*}{60\%} 
          & Drop & 0.695±0.053 & 0.722±0.051 & 0.658±0.054 & 0.676±0.052 \\
          & Mean & 0.663±0.054 & 0.690±0.052 & 0.626±0.055 & 0.644±0.053 \\
    \cmidrule{2-6}
          & Pix2Pix & 0.804±0.037 & 0.856±0.039 & 0.780±0.038 & 0.809±0.038 \\
          & DS-GAN & 0.811±0.038 & 0.846±0.037 & 0.778±0.038 & 0.773±0.040 \\
          & LDM   & 0.842±0.035 & 0.871±0.034 & 0.789±0.037 & 0.827±0.035 \\
          & PASTA & 0.851±0.035 & 0.880±0.033 & 0.791±0.035 & 0.829±0.033 \\
          & FICD & 0.839±0.037 & 0.859±0.037 & 0.788±0.035 & 0.831±0.035 \\
          & ACADiff-emb(ours) & 0.870±0.033 & 0.881±0.032 & 0.808±0.037 & 0.836±0.038 \\
          & ACADiff (ours) & \textbf{0.878±0.030} & \textbf{0.883±0.029} & \textbf{0.819±0.036} & \textbf{0.841±0.038} \\
    \midrule
    \multirow{9}[4]{*}{80\%} 
          & Drop & 0.582±0.058 & 0.609±0.056 & 0.545±0.059 & 0.563±0.057 \\
          & Mean & 0.551±0.060 & 0.578±0.058 & 0.514±0.061 & 0.532±0.059 \\
    \cmidrule{2-6}
          & Pix2Pix & 0.718±0.046 & 0.746±0.047 & 0.675±0.048 & 0.668±0.047 \\
          & DS-GAN & 0.724±0.045 & 0.700±0.052 & 0.670±0.048 & 0.657±0.048 \\
          & LDM   & 0.764±0.049 & 0.757±0.049 & 0.683±0.048 & 0.702±0.049 \\
          & PASTA & 0.759±0.048 & 0.754±0.049 & 0.679±0.048 & 0.696±0.046 \\
          & FICD & 0.722±0.051 & 0.739±0.048 & 0.649±0.051 & 0.660±0.052 \\
          & ACADiff-emb(ours) & 0.768±0.049 & 0.757±0.052 & 0.711±0.048 & 0.704±0.047 \\
          & ACADiff (ours) & \textbf{0.775±0.046} & \textbf{0.763±0.046} & \textbf{0.719±0.041} & \textbf{0.713±0.045} \\
    \bottomrule
    \vspace{-30pt}
    \end{tabular}%
  }
  \label{tab:vol_classification}%
\end{table}

\header{Implementation details} The framework compresses brain volumes from $160{\times}180{\times}160$ to $20{\times}22{\times}20$ via pretrained 3D Autoencoder-KL models with frozen decoders. We implement three independent generators: Any→MRI, Any→FDG-PET, and Any→AV45-PET. The denoiser is a volumetric U-Net with GroupNorm and FiLM modulation, optimized via AdamW (lr=$1{\times}10^{-4}$, $T{=}1000$). ACADiff uses GPT-4o's text encoder for clinical prompts, while ACADiff-emb uses learnable embeddings (dim 512) for comparison. During inference, we perform 10-fold Monte Carlo sampling. For classification, we use a 3D DenseNet-121~\cite{solano2020alzheimer} on completed multimodal volumes. Generation metrics are averaged across three generators; classification over 10 independent generations. Experiments are conducted on 4 NVIDIA A100 GPUs.

\header{Results} Fig.~\ref{fig:gen} shows generation performance. ACADiff consistently outperforms all baselines across all metrics (PSNR 27.9, SSIM 0.911, NMI 0.859, MAE 0.014), exceeding the best baseline LDM. The gap between ACADiff and ACADiff-emb ($\sim$1.8 PSNR) validates the benefit of semantic clinical encoding via GPT-4o. Table~\ref{tab:vol_classification} presents AD vs. HC classification under varying missing rates. ACADiff maintains robust performance across all scenarios, achieving 89.4\% accuracy with 20\% missing data (97.2\% of oracle using complete real data). The advantage becomes more pronounced under extreme conditions: at 80\% missing data, ACADiff preserves 77.5\% accuracy while simple imputation fails. Among baselines, LDM performs best (76.4\%) but remains below our approach. The consistent superiority over ACADiff-emb confirms that language model encoding provides meaningful clinical guidance. These results validate the clinical utility of our framework for multimodal brain image completion.

\section{Conclusion} We presented ACADiff, an adaptive clinical-aware diffusion framework for multimodal brain image synthesis in Alzheimer's disease analysis. By integrating adaptive fusion mechanisms, semantic clinical guidance, and specialized generators, our method effectively handles missing modalities while preserving diagnostic information. Experiments on 1,028 ADNI subjects demonstrate superior performance across all missing rates, with ACADiff maintaining 77.5\% accuracy even at 80\% missing data, outperforming existing methods. These results validate the feasibility of our framework: restoring missing modalities via ACADiff produces more complete multimodal representations, leading to improved diagnostic accuracy. The findings highlight the potential of disease-guided diffusion models to achieve clinically faithful and robust cross-modal synthesis for Alzheimer's disease.
\section{Compliance with ethical standards}
This retrospective study used publicly available human subject data from the ADNI database~\cite{mueller2005alzheimer}. Ethical approval was not required per the open access data license.
\section{Acknowledgements}
This work was supported by NIH (R01LM013519, RF1AG077820, R01MH129694, R21AG080425), NSF (IIS-2319451, MRI-2215789), DOE (DE-SC0025801), Alzheimer's Association (AARG-22-972541), Lehigh University (CORE and RIG), and NSF ACCESS (CIS240554).

\bibliographystyle{IEEEbib}
\bibliography{refs}

\end{document}